\documentclass[letterpaper]{article} 
\usepackage{aaai25}  
\usepackage{times}  
\usepackage{helvet}  
\usepackage{courier}  
\usepackage[hyphens]{url}  
\usepackage{graphicx} 
\usepackage{natbib}  
\usepackage{caption} 
\usepackage{algorithm}
\usepackage{algorithmic}
\usepackage{booktabs}
\usepackage{newfloat}
\usepackage{listings}
\DeclareCaptionStyle{ruled}{labelfont=normalfont,labelsep=colon,strut=off} 
\floatstyle{ruled}
\newfloat{listing}{tb}{lst}{}
\floatname{listing}{Listing}
\title{
Multi-OphthaLingua: A Multilingual Benchmark for Assessing and Debiasing LLM Ophthalmological QA in LMICs}
\author{
    David Restrepo\textsuperscript{\rm 1}\equalcontrib,
    Chenwei Wu\textsuperscript{\rm 2}\equalcontrib, 
    Zhengxu Tang\textsuperscript{\rm 2},
    Zitao Shuai\textsuperscript{\rm 2},
    Thao Nguyen Minh Phan\textsuperscript{\rm 4},\\
    Jun-En Ding\textsuperscript{\rm 7},
    Cong-Tinh Dao\textsuperscript{\rm 4},
    Jack Gallifant\textsuperscript{\rm 5},
    Robyn Gayle Dychiao\textsuperscript{\rm 6},
    Jose Carlo Artiaga\textsuperscript{\rm 6},\\
    André Hiroshi Bando\textsuperscript{\rm 3},
    Carolina Pelegrini Barbosa Gracitelli\textsuperscript{\rm 3},
    Vincenz Ferrer\textsuperscript{\rm 6},\\
    Leo Anthony Celi\textsuperscript{\rm 1},
    Danielle Bitterman\textsuperscript{\rm 5},
    Michael G Morley\textsuperscript{\rm 5},
    Luis Filipe Nakayama\textsuperscript{\rm 1}
}
\affiliations{
    \textsuperscript{\rm 1}Massachusetts Institute of Technology, Cambridge, USA\\
    \textsuperscript{\rm 2}University of Michigan - Ann Arbor, Ann Arbor, USA\\
    \textsuperscript{\rm 3}Universidade Federal de São Paulo, São Paulo, Brazile\\
    \textsuperscript{\rm 4}National Yang Ming Chiao Tung University, Taipei, Taiwan\\
    \textsuperscript{\rm 5}Harvard University, Cambridge, USA\\
    \textsuperscript{\rm 6}University of the Philippines, Manila, Philippine\\
    \textsuperscript{\rm 7}Stevens Institute of Technology, Hoboken, USA\\
    davidres@mit.edu\\
    \textsuperscript{\equalcontrib} Co-first authors.\\
}

\usepackage{bibentry}

\begin{document}

\maketitle

\begin{abstract}
Current ophthalmology clinical workflows are plagued by over-referrals, long waits, and complex and heterogeneous medical records. Large language models (LLMs) present a promising solution to automate various procedures such as triaging, preliminary tests like visual acuity assessment, and report summaries. However, LLMs have demonstrated significantly varied performance across different languages in natural language question-answering tasks, potentially exacerbating healthcare disparities in Low and Middle-Income Countries (LMICs). This study introduces the first multilingual ophthalmological question-answering benchmark with manually curated questions parallel across languages, allowing for direct cross-lingual comparisons. Our evaluation of 6 popular LLMs across 7 different languages reveals substantial bias across different languages, highlighting risks for clinical deployment of LLMs in LMICs. Existing debiasing methods such as Translation Chain-of-Thought or Retrieval-augmented generation (RAG) by themselves fall short of closing this performance gap, often failing to improve performance across all languages and lacking specificity for the medical domain. To address this issue, We propose CLARA (\textbf{C}ross-\textbf{L}ingu\textbf{a}l \textbf{R}eflective \textbf{A}gentic system), a novel inference time de-biasing method leveraging retrieval augmented generation and self-verification. Our approach not only improves performance across all languages but also significantly reduces the multilingual bias gap, facilitating equitable LLM application across the globe.
\end{abstract}

\section{Introduction}

Large Language Models (LLMs) such as Llama and GPT families have emerged as a transformative force in the field of artificial intelligence, offering new tools and solutions in healthcare, with ophthalmology presenting a particularly promising field for their applications \cite{NLP_in_oph3}. Characterized by high patient volumes and substantial documentation burdens \cite{NLP_in_oph1,NLP_in_oph4}, ophthalmology could benefit significantly from LLM integration, mainly in regions such as LMICs where specialized personnel are scarce resources\cite{malerbi2022real}. LLMs could potentially automate remote triage \cite{NLP_in_oph6}, clinical decision support \cite{NLP_in_oph1}, patient education \cite{NLP_in_oph7}, and administrative workflows \cite{NLP_in_oph5}. 

\begin{figure}[ht!]
  \centering
  \includegraphics[width=\columnwidth]{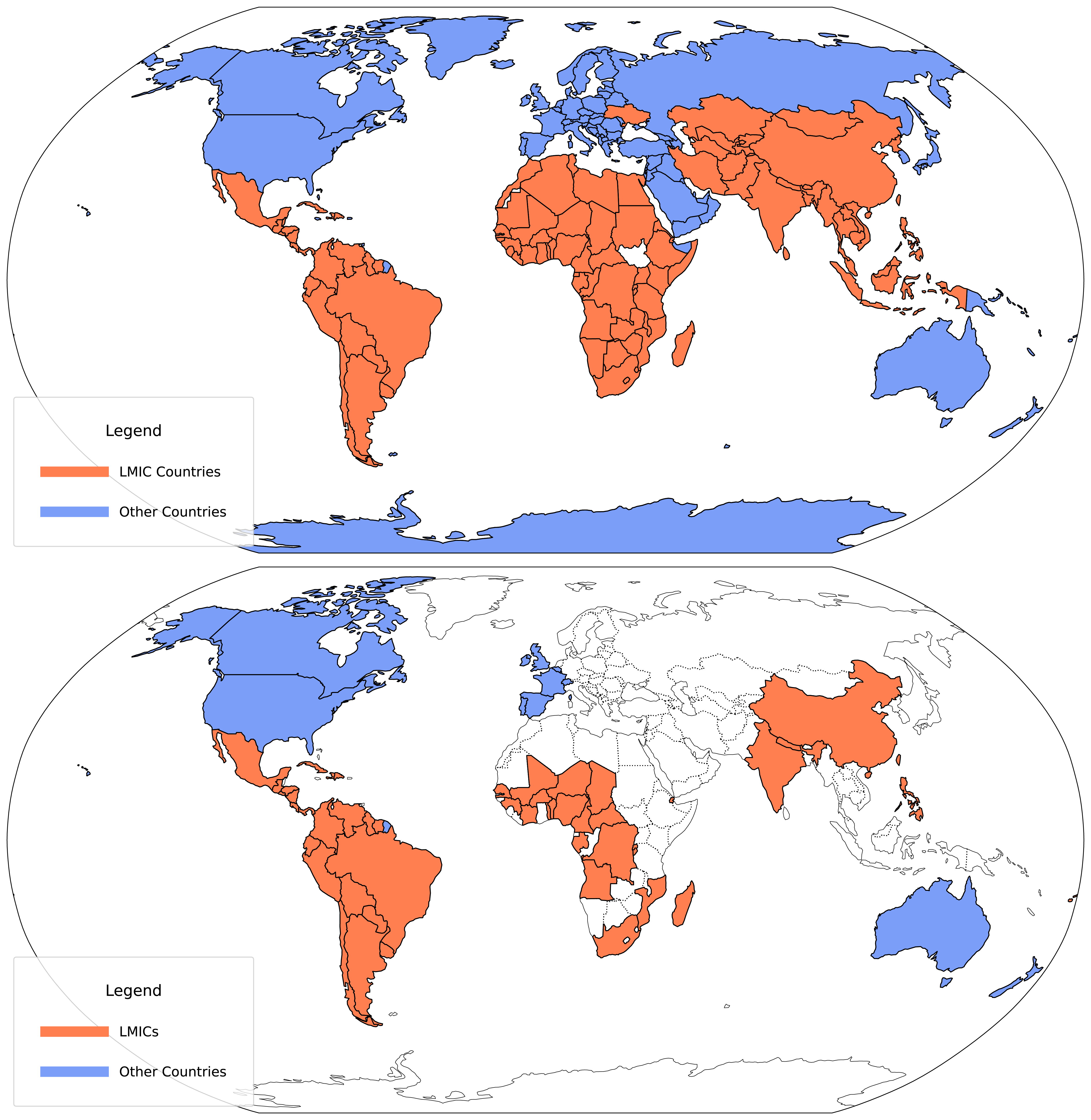}
  \caption{The upper map: Global distribution of Low and Middle-Income Countries (LMICs); The lower map: Countries and regions where the languages included in our benchmark are predominantly spoken (English, Portuguese, Spanish, Filipino, Mandarin, French and Hindi). Our dataset covers 49 LMIC countries and approximately 4.5 billion population.}
  \label{fig:dataset_dist}
\end{figure}

While LLMs have shown remarkable potential in healthcare, their performance is not uniform across all languages \cite{local1,local2,local3}. These models typically perform best in English, benefiting from abundant training data, but often struggle with languages common in Low and Middle-Income Countries (LMICs) such as Portuguese, Spanish, Hindi, and Filipino \cite{local1,local4,local8}, where the amount of ophthalmological data available is considerably limited \cite{2024ophthalmology_bd_review, han2023real}. This disparity poses a significant challenge to the equitable deployment of LLMs in global healthcare \cite{local5,local6,XLT}. Ironically, LMICs may stand to benefit the most from validated LLM applications like remote diagnostics \cite{healthcare3} in the future due to their limited trained personnel \cite{healthcare1,healthcare2}  compared to wealthier nations. Also, effective healthcare delivery in LMICs often requires communication in local languages\cite{healthcare4}. To fully realize the potential of LLMs in improving global health outcomes \cite{bias1}, it's crucial to evaluate and mitigate this linguistic bias \cite{bias2} and ensure these advanced technologies can effectively serve diverse populations \cite{li2024optimizing}, regardless of their primary language \cite{bias3}.

To better understand and address the linguistic disparities in LLM performance for ophthalmology, we present Multi-OphthaLingua, the first multilingual ophthalmological QA dataset, comprised of 1184 questions across English, Spanish, Filipino, Portuguese, Mandarin, French, and Hindi. This dataset, created by ophthalmologists over the globe with diverse language backgrounds, covers a wide range of topics such as basic ophthalmology sciences, clinical cases, surgical practice and post-op care. The questions are structured as multiple-choice with four options each, carefully crafted to ensure question neutrality across regions. Board-certified native-speaker ophthalmologists reviewed each question-answer pair to guarantee robustness in assessing medical knowledge, language understanding, and cultural appropriateness. Most importantly, our dataset features paired questions across all languages, enabling direct cross-linguistic comparisons - a feature absent in most existing multilingual benchmarks, as shown in Table \ref{tab:position}.

\begin{table*}[h] 
\centering
\caption{Comparisons between our proposed Multi-OphthaLingua and related medical QA benchmarks.}
\scalebox{0.75}{
\begin{tabular}{@{} l p{4cm} ll l }
\toprule
\textbf{Dataset} &\textbf{Language} & \textbf{Paired Data}& \textbf{Size} & \textbf{Topic} \\ 
\midrule
MedQAUSMLE~\cite{jin2021disease} & English& No & 11.4k & medical exam QA dataset for disease diagnosis\\
PubMedQA~\cite{jin2019pubmedqa} & English& No & 211k  & biomedical research questions QA dataset\\
MedMCQA~\cite{pal2022medmcqa}  & English& No & 193k  & MCQ dataset for medical entrance exams\\
MMLU-Medical~\cite{hendryckstest2021}  & English& No &116  & multitask medical MCQ \\

MedQA-MCMLE~\cite{jin2021disease} & Mandarin & No & 270k & QA dataset for solving medical problems\\
CMB-single~\cite{cmedbenchmark} & Mandarin & No & 34k & multi-level assessment for medical knowledge\\
CMMLU-Medical~\cite{li2023cmmlu} & Mandarin & No & 242 & Chinese medical assessment suite QA dataset\\
CExam~\cite{liu2023benchmarking} & Mandarin & No & 68k & Chinese medical licensing exam QA dataset \\

Headqa~\cite{vilares-gomez-rodriguez-2019-head} & Spanish & No & 7k& Spanish healthcare exam dataset\\
Frenchmedmcqa~\cite{labrak2023frenchmedmcqa} & French & No & 3k & French medical exam MCQ dataset\\
MMLUHI~\cite{Chen_MultilingualSIFT_Multilingual_Supervised_2023} & Hindi & No & 408 & Hindi medical QA dataset for instruction fine-tuning\\
MMLUAra~\cite{Chen_MultilingualSIFT_Multilingual_Supervised_2023} & Arabic & No & 401 & Arabic medical QA dataset for instruction fine-tuning\\
\textbf{Multi-OphthaLingua (ours)} & English; Spanish; Mandarin; Portuguese; Filipino; Hindi; French& Yes & 1184 & Multilingual Benchmark for Assessing and Debiasing\\
\bottomrule
\end{tabular}
}
\label{tab:position}
\end{table*}

Our evaluation of 6 popular LLMs (Llama-2 70B, Llama-3 70B, Mixtral 8x7B, Qwen-2 72B, GPT 3.5, and GPT 4)  reveals significant disparities in LLM performance across languages, with a notable gap for languages from LMICs such as Filipino. We observed that LLMs generally perform worse on clinical and surgical questions, indicating a lack of specialized domain knowledge. Our qualitative analyses show that these challenges arise from LLMs' limited abilities in LMIC languages, insufficient medical domain knowledge, and struggles with linguistic nuances. Existing work utilizes methods like English-based Chain of Thought (EN-COT) and Pre-Translation \cite{EN-COT} to address language limitations, and Retrieval Augmented Generation (RAG) \cite{rag} to compensate LLMs' lack of medical knowledge. However, these methods face challenges such as low availability of multilingual medical content and quality of matches, especially for LMIC languages. We address these issues by proposing a Cross-Lingual Reflective Agentic system (CLARA) specifically designed for multilingual debiasing in the medical context. CLARA employs a multi-agent approach, including translation, evaluation, knowledge augmentation, and rewriting agents.  Extensive evaluation and ablation demonstrate that our approach simultaneously improves question-answering accuracy across all languages and reduces the cross-lingual performance gap.

The main contributions of this work are three-fold:

\begin{enumerate}
    \item We introduce the first multilingual ophthalmology question-answering benchmark with paired data across seven languages, enabling direct cross-linguistic comparisons. A sample dataset and repo are publicly available at (\url{https://huggingface.co/datasets/AAAIBenchmark/Multi-Opthalingua/tree/main}), and the full version will be released on Physionet, promoting reproducibility and future research in this field. 
    
    \item We provide both quantitative and qualitative investigation of the factors contributing to LLM failures in multilingual ophthalmology contexts, as well as why existing debiasing methods fall short. Our analysis offers valuable insights into the challenges of applying LLMs in diverse linguistic and medical settings.
    
    \item We propose CLARA (Cross-Lingual Reflective Agentic system), a novel inference-time debiasing method specifically tailored for the medical domain. Through extensive evaluation, we demonstrate that CLARA significantly improves both overall accuracy and fairness across languages in ophthalmological question-answering tasks.
\end{enumerate}

\section{Related Work}

\textbf{Existing Work on LLM Multilingual Dataset and Biases in Medicine} LLMs' unequal performance across languages in tasks such as question answering and numerical reasoning has raised concerns, as training corpora are primarily from resource-rich languages \cite{xu2024survey}. Studies have revealed that LLMs in multilingual tasks suffer from language bias, demographic bias, and evaluation bias \cite{xu2024survey}, significantly favoring Western continents in factual information and reasoning \cite{shafayat2024multi, zhao2024gender}. 

There is also a lack of comprehensive multilingual evaluation efforts for medical question-answering with LLMs, particularly in specialized domains like ophthalmology. As shown in Table \ref{tab:position}, existing datasets are predominantly in English or Mandarin, with minimal representation of languages from LMICs. None of the existing datasets provide paired data across multiple languages, which is crucial for consistent evaluation of LLMs' multilingual capabilities. While some datasets are substantial in size (e.g., PubMedQA with 211k entries), they are limited to a single language and often focus on general medical knowledge rather than specialized fields like ophthalmology. This gap underscores the need for a robust benchmark and holistic evaluation.

\textbf{Existing LLM Debiasing Methods for Multilingual QA} Recent efforts have introduced innovative methodologies to enhance the multilingual capabilities of LLMs, primarily in non-medical domains. Notable approaches include Cross-Lingual Template Prompting (XLT) \cite{XLT,rag_2}, Pre-Translate COT \cite{google_multilingual,xcot}, and EN-COT \cite{EN-COT}. XLT \cite{XLT} employs a generic template prompt to engage LLMs' cross-lingual and logical reasoning capabilities. While effective for simple questions, its applicability to complex medical queries may be limited. EN-COT \cite{EN-COT} generates a chain of thought directly in English, regardless of the original problem language, leveraging English for cross-lingual knowledge transfer. Pre-Translate COT \cite{google_multilingual,xcot} involves translating problems into English before applying the English Chain of Thought (CoT) technique. However, it may not fully address the specific linguistic needs of clinicians due to inconsistent translation qualities. Frameworks like causality-guided debiasing\cite{causcal_1}also provide a theoretical grounding for understanding and addressing biases in LLMs. Challenges persist in ensuring these techniques' effectiveness across diverse medical terminologies and contexts.

\textbf{Retrieval Augmented Generation and LLM Self-Verification} While methods like EN-COT\cite{EN-COT} attempt to overcome language limitations, LLMs may still lack domain knowledge and not fully capture the nuances of medical terminology in different languages. Retrieval Augmented Generation (RAG) has emerged as a powerful technique to address LLMs' lack of medical knowledge without requiring extra fine-tuning on the model \cite{llm_healthcare_1,llm_healthcare_2}. However, RAG faces challenges in content relevance and quality of matches, especially for target LMIC languages with limited resources \cite{rag_quality_1,rag_quality_2}. Self-verification techniques have shown potential in enhancing the reliability of AI-generated responses. Frameworks like Self-RAG (Self-Reflective Retrieval-Augmented Generation) \cite{self_rag} use retrieval and self-reflection mechanisms to critique and improve responses iteratively, enhancing the quality and factuality of generated text.

\section{Benchmark Construction, Results and Analysis}\label{sec11}

This section details the processes of data collection, preparation, and technical implementation of our evaluation framework to assess LLM multilingual language understanding and medical knowledge. The proposed dataset aims to address the following key research questions (RQs):

\textbf{RQ1:} Are current LLMs equipped with domain knowledge to be useful for ophthalmologist assistance and safe for patients (e.g. understand clinical terms)? 

\textbf{RQ2:} How and how much does LLM performance vary between languages from HICs and those from LMICs?

\textbf{RQ3:} What factors may contribute to any observed language disparities?

\textbf{RQ4:} Are current debiasing methods adequate to address language disparities?

\subsection{Benchmark Construction}
To construct our dataset, 7 board-certified ophthalmologists with diverse language backgrounds over the globe curated a set of ophthalmological test questions, creating a multilingual dataset of 1184 questions.

The questions were manually designed and inspired by the Brazilian Ophthalmological Board exams, with clinical experts ensuring there would be neutrality in answers across different regions of the world. The benchmark consists of two high-level categories: basic sciences and clinical-surgical ophthalmology, with subgroups on anatomy, pharmacology, clinical optics, strabismus, cataract, uveitis, oncology, refractive surgery, contact lens, and genetics. The distribution of the questions is in Figure \ref{fig:dataset_dist}. 

\begin{figure*}[ht!]
  \centering
  \includegraphics[width=1\textwidth]{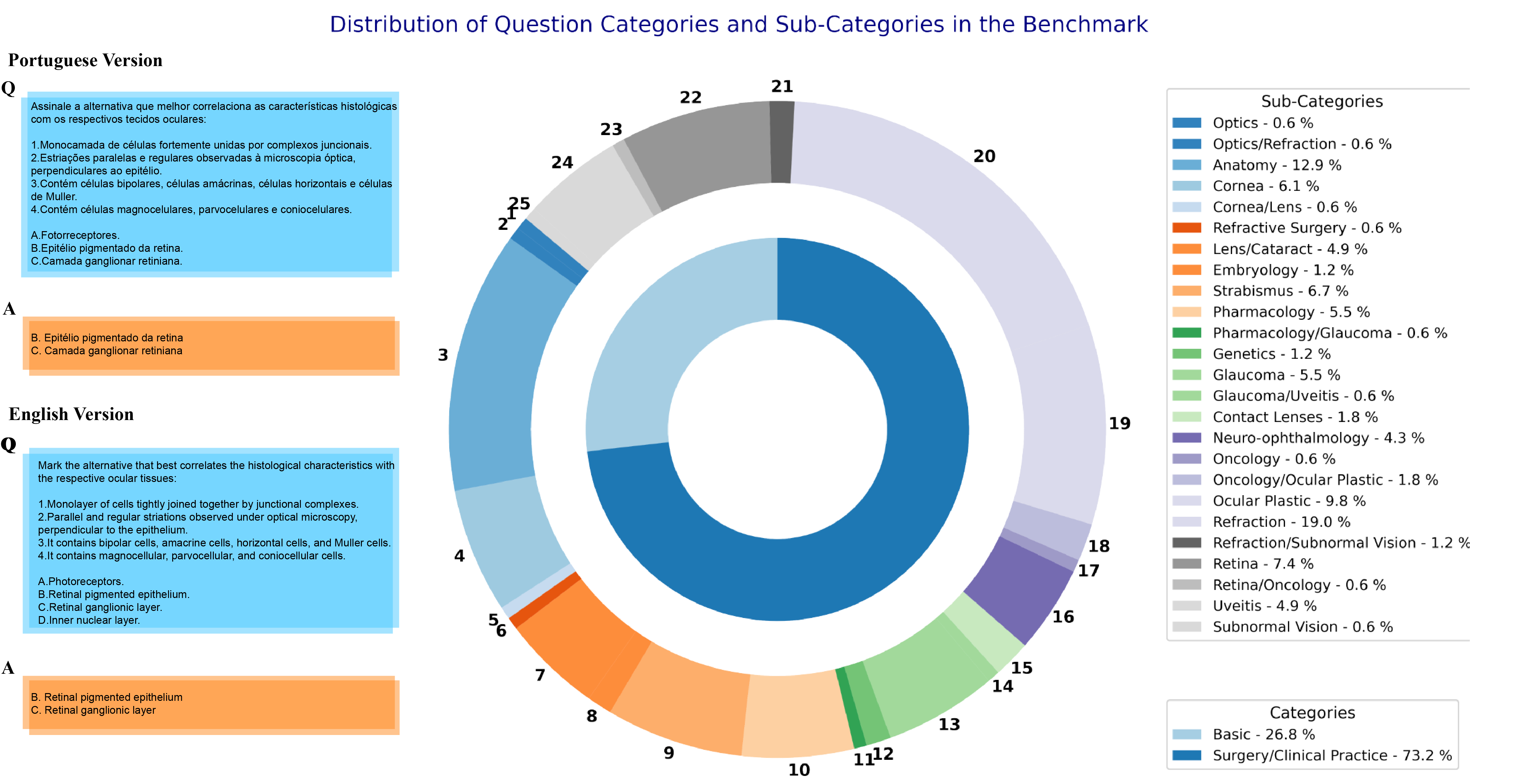}
  \caption{Left: An example question and answer in English and Portuguese; Right: Composition of the benchmark dataset by categories and subtypes.}
  \label{fig:dataset_dist}
\end{figure*}

Originally written in Portuguese, these questions were manually translated into English, Spanish, Filipino, Mandarin, French, and Hindi to ensure broad linguistic, population, and geographical coverage. We select English as the most HIC-representative language, French, Spanish, and Portuguese as languages both spoken in HICs and LMICs, Chinese and Hindi as LMIC languages with large populations, and Filipino as LMIC languages with a small population. Each question-answer pair was reviewed and curated by board-certified native-speaker ophthalmologists to ensure a robust assessment of the models' medical knowledge, language correctness, and cultural appropriateness. A sample question in English and Portuguese is shown in Figure \ref{fig:dataset_dist}.

We evaluated our benchmark against 6 mainstream LLMs. The GPT 3.5 and GPT 4 models were implemented using the OpenAI API in Python 3.9.18, utilizing the OpenAI library version 1.40.3 for interacting with the language models. For the other models (LLAMA 2 70B, LLAMA 3 70B, Qwen-2 72B, and Mixtral 8x7B), Ollama and Together AI library was used. To facilitate a unified evaluation environment and enable advanced prompting techniques\cite{fan2024towards}, we employed the Langchain Python package for prompt template generation and API interaction across all models. We formulated standardized Chain-of-thought prompt templates for question-answering, aiming to eliminate any confounding variables related to differences in prompts or data leakage. The chain-of-thought process was stored for qualitative analysis, providing insights into the models' reasoning patterns. For quantitative evaluation, we implemented a JSON output schema of ABCD choices only and calculated the overall, subdomain-wise, and disease-wise accuracy. Each model's configuration was set to a consistent number to reduce randomness (e.g. temperature = 0). The structure of the prompt is included in the Appendix.

\subsection{Benchmark Results and LLM-Failure Analysis}

\textbf{\textit{The performance of LLMs differs significantly between languages from HICs and those from LMICs.} } As shown in Table \ref{table:llm_perf}, our quantitative evaluation results highlight significant disparities in the LLMs' performance across languages, with the most notable underperformance in Filipino and Hindi. GPT-4 achieved only 51.8\% accuracy in Filipino and 50.6\% in Hindi, compared to 63.4\% in English, a substantial 11.6\% and 12.8 difference respectively. This gap is even more pronounced in Llama-3 70B with a difference of 14\% in Filipino compared to English, and Mixtral, with a difference of 17.6\% for Filipino, and 20.1\% for Mandarin compared to English. As can be seen in table \ref{tab:position}, the tendency of LMIC languages, such as Filipino (FIL), Hindi (HI) and Mandarin (ZH), consistently exhibit similar patterns of decreased accuracy across all models. One exception stands for Mandarin on Qwen-2, potentially due to the larger Mandarin corpus involved in its pretraining and its tokenizer being also trained on Chinese characters \cite{yang2024qwen2}.

\textbf{\textit{LLMs are not equipped with sufficient ophthalmology domain-specific knowledge.} }We observed that more advanced LLMs demonstrate smaller performance gaps between languages. GPT-4, for instance, shows a narrower margin between its highest and lowest-performing languages compared to GPT-3.5. This trend is also evident in open-source models: Qwen-2 72B, which exhibits a smaller gap between English and Filipino performance compared to less sophisticated models like Llama-2 70B or Mixtral-8x7B. However, it is noteworthy that while these models show improvement in reducing language disparities, significant gaps persist. While the concrete composition of every model's pretraining datasets is unknown, it is widely acknowledged that they are largely composed of more common languages, such as English, in online corpora. These findings pinpoint a concerning bias in these models, which tend to favor languages with extensive representation in their training datasets. 

\textbf{\textit{Performance degradation results from difficulties of LLMs in understanding clinic terms, linguistic nuances and cultural context.} }Our analysis also reveals a concerning trend of performance degradation when LLMs encounter more complex, clinically oriented questions. GPT-4's average performance on clinical/surgical questions experiences a 3.79\% drop from its basic science questions, while for Llama-3 and Qwen-2 this number goes up to 4.46\% and 6.84\%. This trend suggests a lack of specialized clinical knowledge in LLMs, which becomes more pronounced when prompting using LMIC languages.

Qualitative analysis by ophthalmologists of the CoT histories revealed striking examples of how LLMs struggle with the nuanced interpretation of medical terminology across different languages, potentially leading to critical misunderstandings in clinical contexts. Our complete qualitative analysis is included in the Appendix. For instance, when presented with the Filipino term ``namamaga," which can mean both ``swollen" and ``inflamed", GPT-3.5 consistently interpreted it as ``swollen" in all contexts, missing the subtle distinction that could be crucial in diagnosing conditions like uveitis. Similarly, in Portuguese, the term ``mancha" can refer to both a ``spot" and a ``stain," leading to ambiguous interpretations in descriptions of retinal abnormalities.

We observed that cultural context significantly influenced the models' responses, particularly in post-operative-related queries. When asked about post-operative care in Spanish, GPT-4 emphasized the role of family support, reflecting cultural norms in many Spanish-speaking countries. However, this emphasis was notably absent in English responses, which focused more on clinical follow-up procedures. This discrepancy highlights the models' inconsistent incorporation of cultural nuances in medicine across languages. In Filipino CoT histories, we noticed a tendency for all models to use more colloquial language when discussing symptoms, often employing idiomatic expressions that lacked the precision required in medical communication. For example, it used  ``parang may lumutang na langaw" (like there's a floating fly) to describe a visual disturbance, an expression that doesn't directly translate to the medical term ``floaters" used in English. 

These qualitative observations highlight not just linguistic barriers but also the need for models to understand and appropriately navigate cultural contexts, idiomatic expressions, and region-specific health concerns across different languages. Such nuanced understanding is needed for the responsible and effective deployment of AI in global healthcare settings.

\begin{table}[ht]
\centering
\caption{Performance Across Different LLMs, Languages and Subgroups, EN-English, ES-Spanish, PT-Portuguese, FIL-Filipino/Tagalog, ZH-Mandarin Chinese, HI-Hindi, FR-French. Results averaged by 8 runs.}
\resizebox{\columnwidth}{!}{%
\begin{tabular}{lccccccc}
\hline
 & \textbf{EN} & \textbf{ES} & \textbf{PT} & \textbf{FIL} & \textbf{ZH} & \textbf{HI} & \textbf{FR} \\
\hline
\textbf{GPT 4} & \textbf{63.4} & \textbf{65.2} & \textbf{61.6} & \textbf{51.8} & 52.4 & 50.6 & \textbf{62.8} \\
\textbf{GPT 3.5} & 46.2 & 38.4 & 39 & 34.8 & 32.3 & 37.2 & 38.4 \\
\textbf{Llama-3 70B} & 57.3 & 51.8 & 53.7 & 43.3 & 46.3 & 44.5 & 53 \\
\textbf{Llama-2 70B} & 33.5 & 32.9 & 29.9 & 26.2 & 29.9 & 29.3 & 33.5 \\
\textbf{Mixtral-8x7B} & 52.4 & 47 & 47 & 34.8 & 32.3 & 42.1 & 48.2 \\
\textbf{Qwen-2 72B} & 58.5 & 56.7 & 55.5 & 49.4 & \textbf{55.5} & 48.8 & 59.1 \\
\hline
\textbf{GPT 4 Basic} & \textbf{65.9} & \textbf{70.5} & \textbf{65.9} & \textbf{45.5} & 59.1 & \textbf{56.8} &\textbf{ 63.6} \\
\textbf{GPT 3.5 Basic} & 47.5 & 39.2 & 41.2 & 38.3 & 38.6 & 38.6 & 40.9 \\
\textbf{Llama-3 70B Basic} & 56.8 & 50 & 61.4 & 43.2 & 54.5 & 45.5 & 61.4 \\
\textbf{Llama-2 70B Basic} & 36.4 & 29.5 & 25 & 25 & 31.8 & 22.7 & 34.1 \\
\textbf{Mixtral-8x7B Basic} & 52.3 & 40.9 & 50 & 27.3 & 50 & 40.9 & 52.3 \\
\textbf{Qwen-2 72B Basic} & 63.6 & 59.1 & 59.1 & 50 & \textbf{63.6 }& 54.5 & \textbf{63.6 }\\
\hline
\textbf{GPT 4 Surgery} & \textbf{62.5} & \textbf{63.3} & \textbf{60} & \textbf{54.2} & 50 & \textbf{48.3} & \textbf{62.5} \\
\textbf{GPT 3.5 Surgery} & 45.7 & 38.1 & 38.2 & 33.5 & 30 & 36.7 & 37.5 \\
\textbf{Llama-3 70B Surgery} & 57.5 & 52.5 & 50.8 & 43.3 & 43.3 & 44.2 & 50 \\
\textbf{Llama-2 70B Surgery} & 32.5 & 34.2 & 31.7 & 26.7 & 29.2 & 31.7 & 33.3 \\
\textbf{Mixtral-8x7B Surgery} & 52.5 & 49.2 & 45.8 & 37.5 & 25.8 & 42.5 & 46.7 \\
\textbf{Qwen-2 72B Surgery} & 53.7 & 51.8 & 54.2 & 49.2 & \textbf{52.5} & 46.7 & 57.5 \\
\hline
\end{tabular}%
}
\label{table:llm_perf}
\end{table}
\begin{table}[ht]
\centering
\caption{Debias Results of GPT-3.5 and GPT-4}
\resizebox{\columnwidth}{!}{%
\begin{tabular}{llccccccc}
\hline
 &  & \textbf{EN} & \textbf{ES} & \textbf{PT} & \textbf{FIL} & \textbf{ZH} & \textbf{HIN} & \textbf{FR} \\
\hline
\textbf{GPT-4} & Direct Inference & 63.4 & 65.2 & 61.6 & 51.8 & 52.4 & 50.6 & 62.8 \\
 & \quad Gap & 0 & 1.8 & -1.8 & -11.6 & -11 & -12.8 & -0.6 \\
 & Web-ToolCall & 68.9 & 65.2 & 65.9 & 56.8 & 54.3 & 59.8 & 64.0 \\
 & \quad Gap & 0 & -3.7 & -3 & -12.1 & -14.6 & -9.1 & -4.9 \\
 & Translate-COT & 69.5 & 67 & 68.2 & 61.6 & 66.2 & 65.3 & 66.7 \\
 & \quad Gap & 0 & -2.5 & -1.3 & -7.9 & -3.3 & -4.2 & -2.8 \\
 & CLARA(Ours) & \textbf{72.2} & \textbf{70.7} & \textbf{71.8} & \textbf{67.1} & \textbf{70.2} & \textbf{68.4} & \textbf{69.8} \\
 & \quad Gap & 0 & -1.5 & -0.4 & -5.1 & -2 & -3.8 & -2.4 \\
\hline
\textbf{GPT-3.5} & Direct Inference & 46.2 & 38.4 & 39 & 34.8 & 32.3 & 37.2 & 38.4 \\
 & \quad Gap & 0 & -7.8 & -7.2 & -11.4 & -13.9 & -9 & -7.8 \\
 & Web-ToolCall & 41.5 & 41.5 & 42.1 & 35.4 & 31.1 & 37.2 & 37.8 \\
 & \quad Gap & 0 & 0 & 0.6 & -6.1 & -10.4 & -4.3 & -3.7 \\
 & Translate-COT & 48.6 & 42.7 & 42.1 & 40.8 & 42.5 & 42.3 & 43.7 \\
 & \quad Gap & 0 & -5.9 & -6.5 & -7.8 & -6.1 & -6.3 & -4.9 \\
 & Ours & \textbf{53.6} & \textbf{48.7} & \textbf{48.8} & \textbf{45.7} & \textbf{48.3} & \textbf{47.8} & \textbf{49.1} \\
 & \quad Gap & 0 & -4.9 & -4.8 & -7.9 & -5.3 & -5.8 & -4.5 \\
\hline
\end{tabular}%
}
\label{table:debias_results}
\end{table}

\begin{figure*}[ht!]
    \centering
    \includegraphics[width=\linewidth]{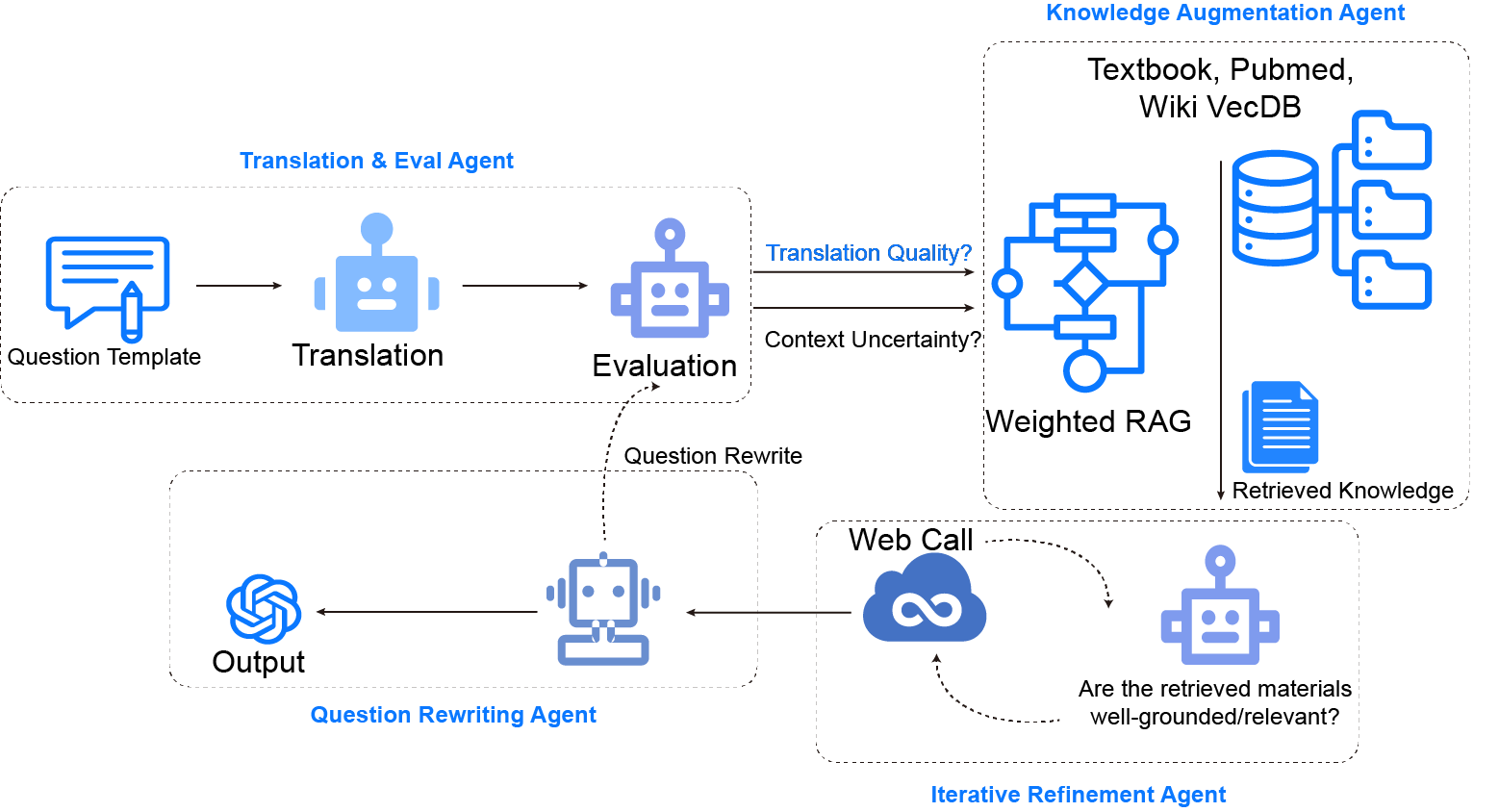}
    \caption{Illustration of CLARA workflow}
    \label{fig:clara_pipelinel}
\end{figure*}

\section{Debiasing LLMs for Multilingual Ophthalmology QA}

Our evaluation of LLMs in multilingual ophthalmology question-answering reveals several existing challenges. Primarily, LLMs exhibit a lack of language proficiency in LMIC languages, compounded by insufficient medical domain knowledge. This knowledge gap often causes models to generate plausible but incorrect reasoning. Another contributor to LLM's poor performance is that LLMs often struggle with understanding nuanced linguistic differences across languages, which can lead to misinterpretations in medical contexts \cite{debias}.

\subsection{Baselines}
Several baseline approaches have been proposed to address these challenges, each with its own limitations. To tackle language differences, methods such as direct inference and pre-translation have been explored. Direct inference attempts to prompt the LLM to process and respond in the target language directly, but this approach may be ineffective for complex medical queries in diverse languages where the model's proficiency is limited. Pre-translation, which involves translating the input to a high-resource language (typically English) and then performing the inference, raises concerns about the accuracy and fidelity of translations, particularly for medical terminology and nuanced symptom descriptions\cite{xcot,google_multilingual}.
Addressing the lack of medical knowledge, some researchers have proposed Web Search Augmentation, utilizing web-search APIs\cite{tao2024harnessing} to ground the LLM's reasoning with a more reliable and traceable knowledge base at inference time\cite{web_1}. However, this method relies heavily on the quality of the knowledge base and the quality of retrieval. Otherwise, this approach risks introducing noisy and incorrect or irrelevant information, potentially leading the LLM to diverge with non-medical or unreliable content. Another approach, RAG, involves retrieving relevant information from curated databases like PubMed or medical textbooks\cite{med_rag}. While promising, RAG faces challenges in the scarcity of high-quality retrieval materials in target LMIC languages, and ensuring the relevance and accuracy of retrieved information remains an active area of research\cite{healthcare1}.

\subsection{CLARA: Cross-LinguAl Reflective Agent}

Inspired by the qualitative and quantitative findings, we propose CLARA (Cross-Lingual Reflective Agentic system), a novel method to address bias in multilingual ophthalmology question-answering. CLARA combines pre-translation, evaluation, corrective RAG, web search, and query transformation within an agentic framework to enhance the robustness and accuracy of responses across diverse languages. Inspired by the concept of agentic AI\cite{llm_agent,llm_agent_2}, CLARA employs multiple specialized agents that collaborate to process and respond to complex queries. Figure \ref{fig:clara_pipelinel} illustrates our pipeline.

\subsubsection{Translation and Initial Evaluation Agent}

The process begins with the Translation Agent converting the input query to English, standardizing inputs across languages. An Evaluation Agent, functioning as a meta-cognitive component, then performs a dual assessment: first, it evaluates the translation quality and certainty; second, it gauges the LLM's overall certainty about the medical question, identifying potential issues with jargon or contextual understanding. This comprehensive evaluation determines the subsequent actions in the pipeline.

\subsubsection{Knowledge Augmentation Agent}

Based on the Evaluation Agent's assessment, CLARA's Knowledge Agent triggers one of the two following RAG approaches. This involves querying a vector embedding database, with weights assigned based on the uncertainty level of different query parts. For parts of the query where translation uncertainty is expressed, a weighted RAG approach is used. We add an additional relevance score to the full-original-query score, here the add-on relevance score $R_i$ for each retrieved document $i$ is calculated as:

\begin{equation}
R_i = \sum(w_j * sim(q_j, d_i))
\end{equation}

where $w_j$ is the weight of query part $j$, and $sim(q_j, d_i)$ is the cosine similarity between query part $j$ and document $i$.

For parts requiring additional context or containing domain-specific jargon, a reweighted RAG approach is applied as well. We perform jargon handling by our Knowledge Augmentation agent, where domain-specific terms are identified and expanded using a specialized ophthalmology dictionary. Key ophthalmological concepts are enriched with relevant contextual information and formatted next to the concept in the prompt. The reweighted relevance score $RR_i$ for each document $i$ is then calculated as:
\begin{equation}
RR_i = \sum(w_j * sim(q_j, d_i)) + \sum(w_k * sim(j_k, d_i))
\end{equation}
where $w_k$ is the weight assigned to jargon term $k$, and $j_k$ is the expanded definition or context for that term. 

We curated a diverse set of RAG corpora from three distinct sources, each contributing unique aspects of medical and general knowledge. PubMed provides a comprehensive repository of biomedical abstracts, widely recognized in the field\cite{2013pubmed}. To ensure in-depth, domain-specific knowledge, we incorporated medical textbooks. Additionally, we included Wikipedia to cover general knowledge aspects to broaden the width of our knowledge base\cite{2014wikidata}. Full details of our RAG sources can be found in Table \ref{table:rag_source}. We use MedCPT-Query-Encoder\cite{medcpt} as our RAG retriever in our model, which is specifically designed for medical domain queries and helps ensure more accurate and relevant retrievals for ophthalmological content.

\begin{table}[h]
\centering
\caption{RAG Corpus Overview}
\begin{tabular}{lcccc}
\toprule
Corpus & \#Doc. & \#Snippets & Length  \\
\midrule
PubMed & 320k & 2.8M & 200 \\
Textbooks & 15 & 106.8k & 200  \\
Wikipedia & 2M & 14.7M & 200 & \\
\bottomrule
\end{tabular}
\label{table:rag_source}
\end{table}
\subsubsection{Iterative Evaluation (of Retrieval) and Refinement Agent}

CLARA employs an iterative evaluation process where a second Evaluation Agent, acting as a critic, assesses the relevance and utility of the retrieved documents. This agent determines whether the retrieved and augmented information is useful for answering the query. If the information is deemed insufficient or irrelevant, it is discarded, and the system's Web Search Tool (using Tavily API) initiates a search for additional information. This iterative process continues until the Evaluation Agent determines that sufficient relevant information has been gathered or a maximum iteration limit = 5 is reached. Then all collected information along with the translated question is passed for final question answering.

\subsubsection{Question Rewriting Agent}

When the maximum rounds are reached, and RAG and web searches fail to yield useful information, CLARA triggers its Rewriting Agent. This situation often arises due to complex, nested information in ophthalmological queries. For instance, consider an ophthalmological examination case that states: ``Bilateral papilledema noted, more pronounced in the right eye with flame-shaped hemorrhages, the left eye showing early cotton wool spots and an area of possible choroidal neovascularization temporal to the macula." Here, the right eye's information is omitted and may cause confusion. Our Rewriting Agent re-composes the translated question and passes the rewritten question back to the first evaluation agent, addressing the limitation in complex information extraction from the query. This structured rewriting helps the LLM to process intricate ophthalmological queries and allows for more targeted information retrieval in subsequent RAG iterations.

Through this multi-agent, reflective process, CLARA aims to mitigate biases and enhance accuracy in multilingual ophthalmology QA, addressing the challenges of language diversity and domain-specific knowledge in medical AI.

\subsection{Debias Results Analysis and Ablation Studies.}
\begin{table}[ht]
\centering
\caption{Ablation Study for GPT-4, 1:Translate-COT, 2:Web ToolCall, 3:Basic-RAG, 4:Corrective-RAG, 5:CLARA}
\resizebox{\columnwidth}{!}{%
\begin{tabular}{cccccc|ccccccc}
\hline
\textbf{1} & \textbf{2} & \textbf{3} & \textbf{4} & \textbf{5} & & \textbf{EN} & \textbf{ES} & \textbf{PT} & \textbf{FIL} & \textbf{ZH} & \textbf{HIN} & \textbf{FR} \\
\hline
N & N & N & N & N & & 66.4 & 63.2 & 63.4 & 51.8 & 52.4 & 50.6 & 62.8 \\
Y & N & N & N & N & & 69.5 & 67.0 & 68.2 & 61.6 & 66.2 & 65.3 & 66.7 \\
Y & Y & N & N & N & & 69.6 & 67.5 & 68.4 & 61.8 & 66.4 & 65.5 & 67.1 \\
Y & Y & Y & N & N & & 70.1 & 68.9 & 69.3 & 62.5 & 68.2 & 67.4 & 68.7 \\
Y & Y & Y & Y & N & & 70.8 & 69.4 & 69.7 & 64.9 & 69.3 & 67.9 & 69.5 \\
Y & Y & Y & Y & Y & & \textbf{72.2} & \textbf{70.7} & \textbf{71.8} & \textbf{67.1} & \textbf{70.2} & \textbf{68.4} & \textbf{69.8} \\
\hline
\end{tabular}%
}
\end{table}
\textbf{\textit{Current de-bias methods are not sufficient to address the performance disparity between HICs and LMICs languages.} }The results in Table \ref{table:debias_results} demonstrate the varied effectiveness of debiasing techniques for multilingual ophthalmology question-answering using GPT-3.5 and GPT-4, revealing important insights when analyzing both performance gaps (relative to English) and absolute accuracy. Direct Inference, the baseline approach, exhibits the largest performance gaps between English and LMIC languages, particularly for Filipino (11.6\% for GPT-4, 11.4\% for GPT-3.5), Mandarin (11.0\% for GPT-4, 13.9\% for GPT-3.5), and Hindi (12.8\% for GPT-4, 9.0\% for GPT-3.5).  This method's overall accuracy is also the lowest across languages, leaving lower-resource languages at a significant disadvantage. 

Web-ToolCall demonstrates inconsistent results, both in terms of fairness gap reduction and absolute performance. For GPT-3.5, it hurts the performance of French and Mandarin by 0.6\% and 1.2\% and offers no value for Hindi. For GPT-4, while it improves absolute performance for most languages, we observe a concerning larger performance gap for Spanish, Portuguese, and French. This inconsistency highlights the unreliability of Web-ToolCall, suggesting that web search processes reliant on scarce LMIC medical text may introduce noisy information, particularly for non-English queries. Translate-COT \cite{TRANSLATE,EN-COT} shows more consistent improvements in both gap reduction and absolute performance across languages for both models. Notably, for GPT-4, it helps Mandarin, Hindi and Filipino pass the benchmark (60\%). However, it still falls short of achieving parity for Hindi and Filipino, especially for less advanced LLMs like GPT-3.5.

Our proposed method demonstrates superior performance in both aspects. It consistently reduces performance gaps across all languages while simultaneously improving absolute accuracy, even for English. Compared to Translate-COT, our method not only narrows the Filipino-English gap from 7.9\% to 5.1\% but also boosts Filipino performance from 61.6\% to 67.1\%. Similar improvements are seen in Mandarin and Hindi.

The ablation study for GPT-4 reveals the incremental benefits of each component in CLARA. Details of how each component is implemented is included in Appendix. Translation alone yields substantial improvements, particularly for lower-resource languages, with Filipino, Mandarin, and Hindi seeing increases of 9.8\%, 13.8\%, and 14.7\%, respectively. Web search capabilities add modest but consistent gains across languages when coupled with Translation, indicating that English-based web retrieval is more reliable than LMIC language-based. Basic RAG further enhances performance, with Filipino, Chinese, and Hindi improving by an additional 0.7\%, 1.8\%, and 1.9\%. The corrective-RAG component shows language-dependent improvements, most notably for Filipino (+2.4\%). The full CLARA system achieves the best results, with Filipino, Mandarin, and Hindi reaching 67.1\%, 70.2\%, and 68.4\%. Importantly, CLARA also improves English performance from 66.4\% to 72.2\%, indicating enhanced overall model capability. 
\section{Conclusion and Broader Impacts}
Our work contributes to more equitable access to medical information across diverse linguistic communities, potentially enhancing healthcare delivery in resource-limited settings. While our method improves performance across languages, the persistent limitations in less advanced models like GPT-3.5 raise ethical concerns about deploying medical AI systems in the real world. This underscores the need for continued research and careful consideration of AI's role in healthcare.


\end{document}